# Emulating On-Orbit Interactions Using Forward Dynamics Based Cartesian Motion

**Mohatashem Reyaz Makhdoomi**   Doctoral Candidate, Space Robotics Research Group (SpaceR), Interdisciplinary Centre for Security, Reliability and Trust (SnT), University of Luxembourg, Luxembourg. mohatashem.makhdoomi@uni.lu

**Vivek Muralidharan**   Research Associate, Space Robotics Research Group (SpaceR), Interdisciplinary Centre for Security, Reliability and Trust (SnT), University of Luxembourg, Luxembourg. vivek.muralidharan@uni.lu

**Kuldeep R. Barad**   Doctoral Candidate, Space Robotics Research Group (SpaceR), Interdisciplinary Centre for Security, Reliability and Trust (SnT), University of Luxembourg, Luxembourg. kuldeep.barad@uni.lu

**Juan Sandoval**   Associate Professor, Laboratoire des Sciences du Numérique de Nantes : LS2N, Ecole Centrale de Nantes, 44321 Nantes, France. juan.sandoval@ec-nantes.fr

**Miguel Olivares-Mendez**   Assistant Professor, Space Robotics Research Group (SpaceR), Interdisciplinary Centre for Security, Reliability and Trust (SnT), University of Luxembourg, Luxembourg. miguel.olivaresmendez@uni.lu

**Carol Martinez**   Research Scientist, Space Robotics Research Group (SpaceR), Interdisciplinary Centre for Security, Reliability and Trust (SnT), University of Luxembourg, Luxembourg. carol.martinezluna@uni.lu

*ABSTRACT*

On-orbit operations such as servicing and assembly are considered a priority for the future space industry. Ground-based facilities that emulate on-orbit interactions are key tools for developing and testing space technology. This paper presents a control framework to emulate on-orbit operations using on-ground robotic manipulators. It combines Virtual Forward Dynamics Models (VFDM) for Cartesian motion control of robotic manipulators with an Orbital Dynamics Simulator (ODS) based on the Clohessy Wiltshire (CW) Model. The VFDM-based Inverse Kinematics (IK) solver is known to have better motion tracking, path accuracy, and solver convergency than traditional IK solvers. Thus, it provides a stable Cartesian motion for manipulators based on orbit emulations, even at singular or near singular configurations. The framework is tested at the ZeroG-Lab robotic facility of the SnT by emulating two scenarios: free-floating satellite motion and free-floating interaction (collision). Results show fidelity between the simulated motion commanded by the ODS and the one executed by the robot-mounted mockups.

**Keywords:** Virtual Robotic Models, Forward Dynamics, Inverse Kinematics, Orbital Motion, Clohessy-Wiltshire (CW)

# 1 Introduction

Reliable proximity operations on orbit are critical for modern space applications like on-orbit servicing, assembly, and manufacturing. Proximity operations are characterized by a safe approach under constraints followed by the subsequent interaction and manipulation of target objects. Due to complex contact dynamics and sensitive responses in a micro-gravity environment, the on-ground verification of system behavior, reliability, and safety in various on-orbit operations is essential. The latter is addressed using reliable ground-based testing setups like the one shown in Fig. 1, where one of the critical challenges is to reproduce multi-body kinematics, dynamics, and interaction in micro-gravity environments. High-fidelity numerical models may be used to test and analyze the effect of multi-body motion and interaction. However, they tend to ignore the non-deterministic artifacts of a physical system. Physical simulations may also be conducted with full-scale space hardware, although it is often intractable, cost-ineffective, and unnecessary. On the other hand, Hardware-in-the-loop (HIL) testing provides a hybrid approach where the subsystem elements or a subset of all system components may be utilized while other parts of the problem are appropriately ignored or compensated with accurate numerical simulation.

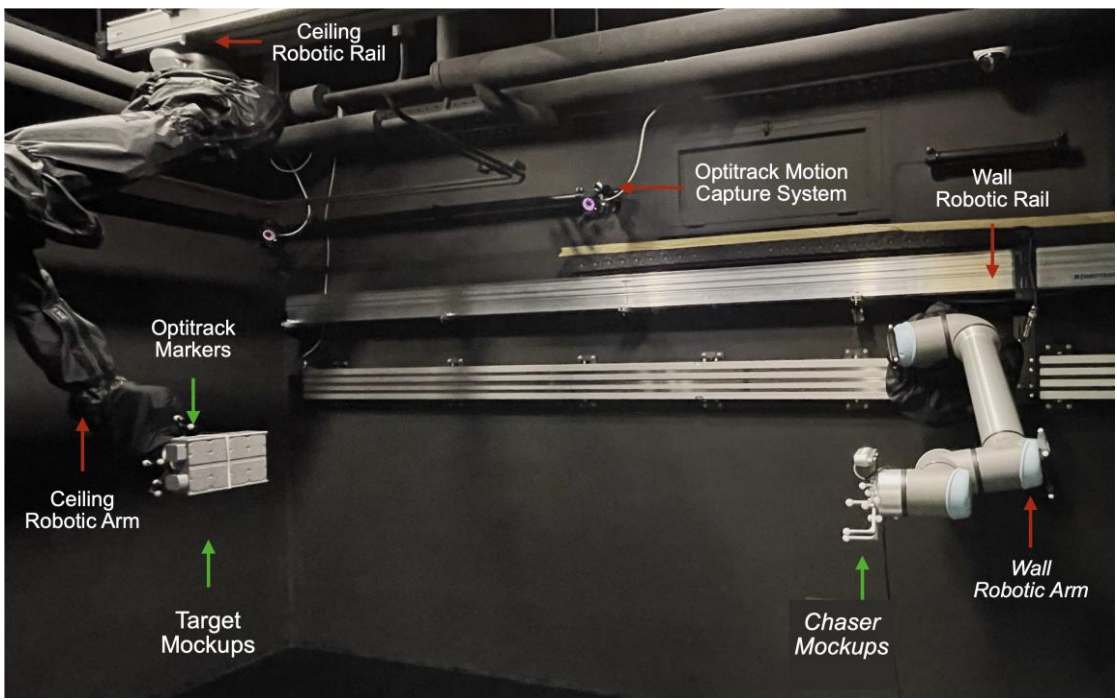

**Fig. 1 The Zero-G lab facility at the University of Luxembourg. A multi-purpose testing facility for emulating on-orbit operations**

Ground facilities that enable HIL testing are important for space systems development. Typically, they enable the recreation of different orbital maneuvers (of up to 6-DOF motion in Cartesian space), using one or more linear tracks, gimbals, robotic manipulators, and suspension cables. For instance, NASA's rendezvous docking simulator that supported verification of docking and proximity operations for Gemini and Apollo missions used suspension cables to allow translation motion and a suspended gimbal to allow rotation [1]. Large-scale HIL facilities have enabled consistent advancement in rendezvous and proximity operations for the International Space Station, where large full-scale satellites are tested in representative visual and dynamic conditions. Examples of this include NASA's Docking overhead target simulator [2], the JAXA RDOTS based on the Cartesian motion table, Lockheed Space Operations Simulation Center (SOSC), and DLR's European Proximity Operations Simulator (EPOS-1). While these facilities enable 6-DOF simulation, it is challenging to isolate the dynamics of the ground environment and the testbed apparatus. The latter is essential when testing actuators or force-based interaction. It can be challenging to reproduce accurate dynamic responses in a multi-body system. For accurate reproduction of dynamic responses, air-bearing setups [3], drop towers and parabolic flights [4], neutral buoyancy [5] and even orbital testbeds [6] may be used for this purpose.

Robotic manipulators form an essential component of most HIL testing facilities. The accuracy of HIL simulations relies on the accuracy with which a manipulator can execute its motion in Cartesian space. In this regard, identifying a solution to the robot's inverse kinematics (IK) problem is essential for close-to-real HIL emulations. Several implementations of Inverse Kinematics exist in the literature, including methods based on Jacobian pseudo-inverse [7], which are accurate but unstable at singularities; Jacobian transpose methods, which are stable [8] but require dynamic decoupling; and optimization methods [9] that are robust but computationally expensive.

Within the purview of ground-based HIL testing facilities for on-orbit scenarios, limited literature discusses the choice of IK solvers for emulating such scenarios with robotic manipulators. Most of the research uses standard IK solvers for robotic motion tasks (provided by the robot manufacturer) while focusing on on-orbit interaction and trajectory planning [10–12]. A typical assumption when working with robotic manipulators is that they are not in singular or near-singular configurations when performing IK calculations. The latter can be a problem during on-orbit emulations, as the manipulator may reach a near-singular configuration while trying to emulate the simulated motion provided by the orbital dynamics simulator (ODS), and thus may behave unstably. Moreover, when emulating on-orbit scenarios, to ensure high fidelity, IK solvers must accurately deliver a solution to sparsely sampled target poses in the Cartesian space provided by the ODS. These discrete targets represent a challenge to controlling the robot's motion since the solution should simultaneously ensure joint space coherency and task space feasibility. To solve the challenges mentioned above, we propose to use a forward dynamics approach to Cartesian Motion Control based on a virtual robot model. This approach introduces a conditioned mass matrix that assists in a smooth transition between sparse targets for accurate motion tracking[13]; and the idea of decoupling virtual dynamics, enabling stable behavior at singular configurations [14].

Therefore, this paper explores the potential of the Virtual Forward Dynamics Models (VFDM) for Cartesian Motion Control of robotic manipulators in space-related applications. The contribution is two-fold 1) We extend the use of VFDM for Cartesian Motion Control of two robotic manipulators to emulate on-orbit scenarios in a multi-purpose ground-based HIL testing facility. 2) We propose a control framework that integrates an Orbital Dynamic Simulator (based on the Clohessy-Wiltshire model) with the VFDM-based Cartesian Motion Control framework. To our knowledge, the utilization of VFDM for emulating on-orbit scenarios has not been previously explored. Compared to traditional IK methods, VFDM is inherently stable and accurate at singular configurations; and ensures smooth robotic motion throughout the workspace, which is paramount for emulating on-orbit trajectories. Experiments are performed using the two UR10e robots of the ZeroG-Lab Hardware-in-the-Loop (HIL) testing facility at the University of Luxembourg, where a wide range of on-orbit scenarios can be emulated [15, 16]. In this paper, we focus on two scenarios: free-floating behavior and free-floating interaction (such as a collision between two satellites), to demonstrate the viability of the framework to emulate interactions in a simulated space environment.

## 2 Orbital Dynamics Simulator (ODS)

Using Kepler's laws of planetary motion, a time-invariant two-body dynamics defines the satellite motion in this investigation. A free-floating motion of a satellite relative to an inertial frame centered on the Earth is not feasible for simulating space environments in a ground-based test facility. The same is valid for observing interactions between two satellites in the inertial frame. Such a challenge is overcome by observing the satellite motion from the perspective of a virtual observer in a rotating frame $\underline{R}$ as demonstrated in Fig. 2. The satellite of interest may be in an elliptical orbit, but with an assumption that the virtual observer is in a nearby circular orbit, a simplified model of relative orbital motion is approximated by the Clohessy-Wiltshire (CW) model [17].

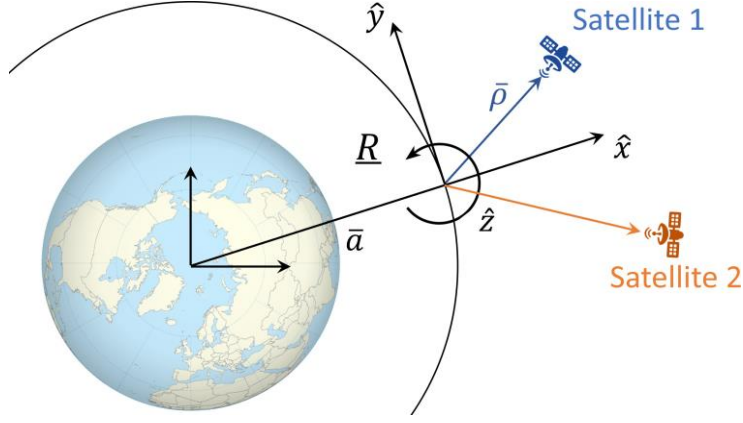

**Fig. 2 Clohessy-Wiltshire model with a virtual observer defined at the origin of the rotating coordinate frame, $\underline{R}$. Direction $\hat{x}$ points radially away from the Earth, $\hat{y}$ points in the direction of orbit velocity and $\hat{z}$ is in the direction of the positive angular momentum vector.**

The motion of the satellite is defined by vector $\bar{\rho}$ relative to the $\underline{R}$ frame, such that $\bar{\rho} = [x, y, z]^T$ [18]. For any external forces $F_1$, $F_2$ and $F_3$ that a satellite may experience along $\hat{x}$, $\hat{y}$, and $\hat{z}$ directions, respectively, the motion is defined by

$$\ddot{x} = 3\Omega^2 x + 2\Omega \dot{y} + \frac{F_1}{m} \qquad (1)$$

$$\ddot{y} = -2\Omega \dot{x} + \frac{F_2}{m} \qquad (2)$$

$$\ddot{z} = -\Omega^2 z + \frac{F_3}{m} \qquad (3)$$

where the quantity $m$ is the mass of the satellite and $\Omega = \sqrt{\mu/a^3}$ is the orbital angular velocity for the virtual observer in an orbit with radius $a$, and $\mu$ is the standard gravitational parameter for Earth. The CW model also accounts for the centrifugal force experienced by the satellites in orbit as well as the Coriolis force when the observer is in the rotating frame of reference. Similarly, the satellites' approximate attitude dynamics are modeled so that the natural gravity torque due to the Earth and any external torques influence the orientation [18]. The angular velocities, $\omega_i$, and orientation quaternions, $\epsilon_i$, evolve as

$$\dot{\omega}_1 = \frac{I_3 - I_2}{I_1}\left(3C_{12}C_{13}\Omega^2 - \omega_2\omega_3\right) + \frac{T_1}{I_1} \qquad (4)$$

$$\dot{\omega}_2 = \frac{I_1 - I_3}{I_2}\left(3C_{11}C_{13}\Omega^2 - \omega_1\omega_3\right) + \frac{T_2}{I_2} \qquad (5)$$

$$\dot{\omega}_3 = \frac{I_2 - I_1}{I_3}\left(3C_{11}C_{12}\Omega^2 - \omega_1\omega_2\right) + \frac{T_3}{I_3} \qquad (6)$$

$$\begin{bmatrix} \dot{\epsilon}_x \\ \dot{\epsilon}_y \\ \dot{\epsilon}_z \\ \dot{\epsilon}_w \end{bmatrix} = \frac{1}{2}\begin{bmatrix} \epsilon_w & -\epsilon_z & \epsilon_y & \epsilon_x \\ \epsilon_z & \epsilon_w & -\epsilon_x & \epsilon_y \\ -\epsilon_y & \epsilon_x & \epsilon_w & \epsilon_z \\ -\epsilon_x & -\epsilon_y & -\epsilon_z & \epsilon_w \end{bmatrix}\begin{bmatrix} \omega_1 \\ \omega_2 \\ \omega_3 \\ 0 \end{bmatrix} \qquad (7)$$

where C is the rotation matrix corresponding to the orientation of the satellite body frame relative to the rotating coordinate frame $\underline{R}$. The quantities $\omega_i$ and $\epsilon_i$ indicate the evolution of the satellite's body relative to the inertial frame. Moreover, $I_1$, $I_2$ and $I_3$ are the moments of inertia along the principal axes of the satellite's body. Finally, $T_1$, $T_2$ and $T_3$ are external torques acting on the satellite.

The Orbital Dynamic Simulator (ODS) uses the aforementioned dynamics to model one or more satellites in the neighborhood. A range of satellite motions can be achieved with specific initial configurations, including gravity-stabilized orientation, rendezvous, collisions, interactions, and free-floating and free-flying behavior.

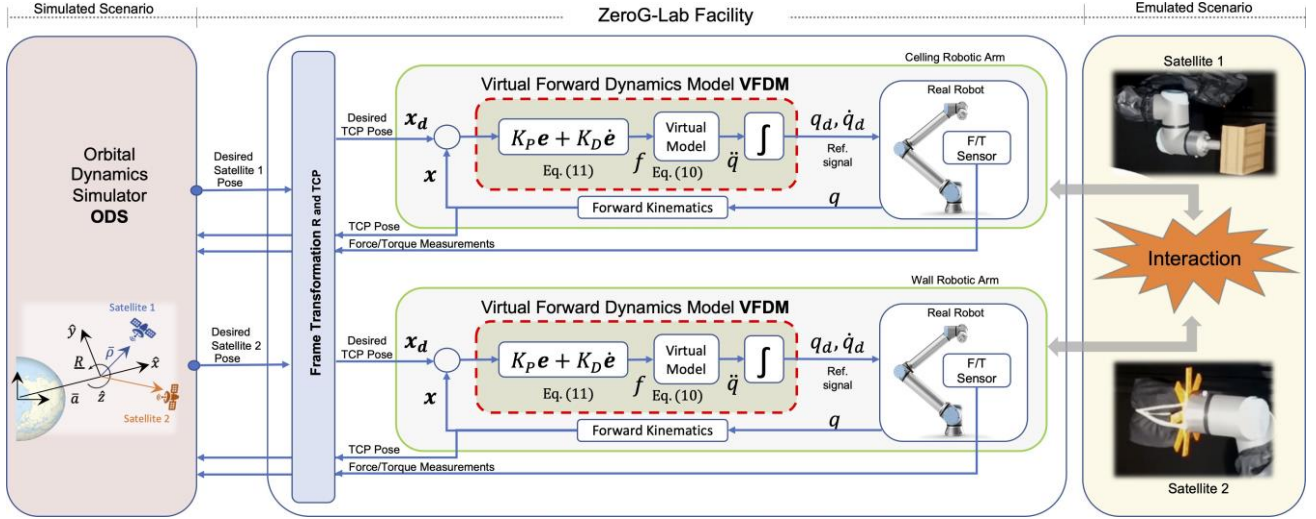

**Fig. 3 Block Diagram for emulating on-orbit scenarios. The Orbital Dynamics Simulator (ODS) imparts the motion that the robots must execute to emulate the motion of satellites in orbit, based on on-the-fly force/torque measurements and state measurements. Red dashed rectangles represent the virtual forward dynamics models. Green rounded rectangles highlight the robot control block.**

## 3 Virtual Forward Dynamics based Cartesian Motion Control

A forward dynamics model describes the effect of forces and torques on the motion of a body. Consider a "virtual" robotic manipulator with identical kinematics as the real robot, described by the following dynamic equation:

$$\tau = H(q)\ddot{q} + C(q, \dot{q}) + G(q) \qquad (8)$$

where the positive definite inertia matrix is denoted by $H(q)$; $C(q, \dot{q})$ describes the effect of Coriolis and Centrifugal forces; $G(q)$ exemplifies the effect of gravity; $\tau$ represents the torques in the joints; and finally, $\ddot{q}$, $\dot{q}$, $q$ are the joint space accelerations, velocities, and positions, respectively. The mapping between end-effector force $f$ and joint torques $\tau$ is given by

$$\tau = J^T f \qquad (9)$$

where $J$ is the Jacobian of the virtual model. Assuming that the virtual model does not need to account for gravity, and needs to accelerate from zero in each control cycle, then the terms $G(q)$ and $C(q, \dot{q})$ are neglected. An important remark here is that the Coriolis and Centrifugal terms of the manipulator in Eq. (8) are not to be confused with the Coriolis and Centrifugal terms accounted for by the CW equations presented in Section 2. Rearranging Eq. (8) to solve for acceleration and omitting the notation $q$ from $H(q)$ for brevity, we have:

$$\ddot{q} = H^{-1} J^T f \qquad (10)$$

consequently, a simplified virtual model, independent of the non-linear joint velocities, described by Eq. (10) relates the Cartesian space forces to instantaneous joint space accelerations. Such a simplification paves the way for conditioning of the joint-space inertia matrix $H$ where a manipulator's dynamic behavior is modified to obtain any desired effects by changing the mass distribution of virtual links. It is through this conditioning that Cartesian Motion Control can be implemented providing both accurate motion tracking and smooth transition between sparsely sampled poses [13]. Additionally, Eq. (10) resembling a Jacobian transpose IK method is rendered stable by dynamic decoupling by considering:

$$\dot{x} = J\dot{q} \quad \Rightarrow \quad \ddot{x} = \dot{J}\dot{q} + J\ddot{q} \tag{11}$$

where assuming the discrete motion for each cycle starts at rest, the term $\dot{J}\dot{q} = 0$. Substituting Eq. (11) in Eq. (10) yields

$$\ddot{x} = JH^{-1}J^T f = M^{-1} f \tag{12}$$

where the quantity $JH^{-1}J^T$ is the inverse of the virtual model's operational space inertia matrix $M$. Through dynamic decoupling $M^{-1}$ is made a time-invariant diagonal matrix across joint configurations that ensures the stability of the model at singular configurations [14]. In Eq. (12), the mapping between forces and accelerations in Cartesian space is decoupled, implying that, ideally, if a force is applied to the end-effector of the virtual model along z-axis of the end-effector frame, for instance, the end-effector should accelerate along z-axis only. Such behavior is then imposed on the real robot. Formulating a closed-loop control around this virtual model, and defining the error term as $\mathbf{e} = x_d - x$, i.e., the difference between the desired target $x_d$ and current end-effector pose $x$ obtained via a forward kinematics (FK) routine. Then, the control input $f$ to the forward model is given by

$$f = K_P \mathbf{e} + K_D \dot{\mathbf{e}} \tag{13}$$

which is expressed as a proportional derivative PD control law. Here $\mathbf{e}$ is a vector with translational and rotational components. The overall framework to emulate on-orbit scenarios is illustrated in a block diagram in Fig. 3. The middle block summarizes the VFDM based Cartesian Motion Control as implemented in the ZeroG-Lab facility. Within the middle block are two robot control blocks, each constituting a UR10e robot equipped with an F/T sensor, controlled via a position interface by a VFDM based Cartesian controller. Each red-dashed rectangle depicts the VFDM model consisting of a virtual robot model described by Eq. (10) and a PD controller Eq. (13). The virtual forward dynamics model (see red dashed rectangle(s) serves as an IK solver and obtains inputs in the form of the satellite motion generated by the Orbital Dynamics Simulator (ODS) Block. The output of this solver (i.e. joint positions) serves as set-points to the robot's internal position controller. The Orbital Dynamics Simulator (ODS) Block is responsible for simulating the dynamics of satellites using the Clohessy-Wiltshire (CW) model and generating satellite motion way-points via a Frame-Transformation Block. The Frame Transformation block expresses this information (pose, force/torque values) in the appropriate frame and exchanges data both ways. The robot control block takes the simulated (or desired) waypoints from the ODS Block and imparts that motion to a mockup mounted on the robot flange. Any mockup mounted on the robots constitutes the emulated scenario, and their states i.e., pose and force/torque information, etc, are obtained via F/T sensors mounted on the robot.

## 4 Experimental Setup

The control framework for emulating on-orbit scenarios proposed in this paper is tested at the ZeroG-Lab at the University of Luxembourg[19]. Two satellite mockups are mounted on the UR10e robot end-effectors as demonstrated in Fig. 4. The inertial properties and the assumed orbital parameters of the mockups are specified in the orbital dynamics simulator (ODS). Upon contact, the forces and moments acting on the mockups are measured by inbuilt F/T sensors of the robot, based on which the on-orbit simulator, through proper force/torque transformations, generates trajectories on the fly. The VFDM-based Cartesian motion control algorithm [20] for controlling the robots and the orbital dynamic simulator (ODS) operate as two ROS packages running in two separate machines (control and simulator computers). Rails are not active and the motion provided by the ODS is executed using the robots only. For this paper, two scenarios were emulated. However, other scenarios such as gravity-stabilized orientation, free-flying behavior, and rendezvous are also possible [16, 21].

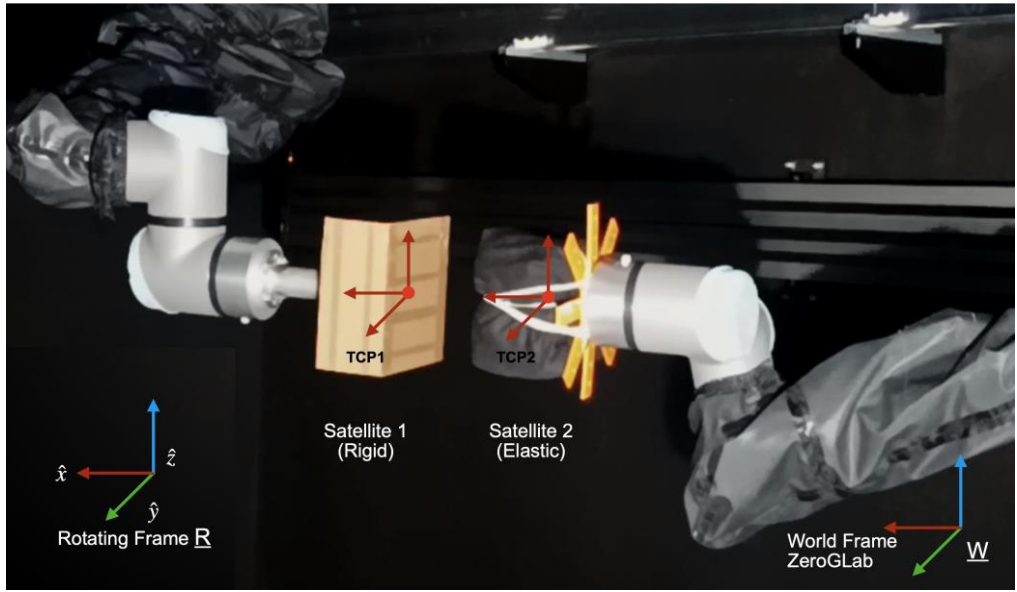

**Fig. 4 ZeroG-Lab Setup. Mockup satellites were mounted at the flange of both robotic arms for the on-orbit emulation tests.**

- **Scenario 1: Free-floating Satellite Motion**

For this scenario, assume the presence of only Satellite 1 in Fig. 2 near an observer in a low earth orbit (LEO) at an altitude of 800 km from the Earth's surface. The motion of Satellite 1 is expressed relative to the rotating coordinate frame $\underline{R}$. For convenience, the initial velocity of the satellite relative to $\underline{R}$ is considered zero. It is assumed that no net generalized forces expressed in $\underline{R}$ are acting on the satellite. Any change in the satellite states (velocities or forces) means the satellite moves in a free-floating motion with respect to $\underline{R}$. To demonstrate the free-floating behavior, external forces are applied, resulting in the satellite accelerating in the direction of the applied forces. In the ZeroG-lab facility, Satellite 1 shown in Fig. 4 is pushed by a human operator in various directions at different instances in time. The F/T sensor measures the forces applied. The forces are then expressed in the $\underline{R}$ frame and supplied as input to the ODS which generates the motion imparted to the mockup. Consequently, the mockup moves as if it were actually free-floating in micro-gravity. The rigid mockup ensures that a reproducible motion is achieved if the same sequence of forces were to be imparted.

- **Scenario 2: Free-Floating Interaction (Collision)**

An interaction between two satellites is emulated by considering the simultaneous existence of Satellites 1 and 2 as shown in Fig. 4. From a dynamics perspective, both orbital docking and collision are comparable. In orbital docking, typically one or more bodies in consideration have control capabilities, and the relative velocity during contact is close to zero. While for collision it is considered that these bodies do not have control capabilities and have non-zero velocities during the contact. In this investigation, a controlled collision is considered to exemplify the interaction, especially because the impact of force transfer is more pronounced. The density of space debris is exceptionally high at an altitude of 800 km [22]. A collision between satellites is portrayed by placing a rotating observer in a circular low earth orbit (LEO) at an altitude of 800 km from the surface of the Earth. The satellites are placed sufficiently close to each other and opposing initial velocities are deliberately imparted to generate a collision. External forces are absent on both the satellite mockups at the initial time. Satellite 1 is a rigid mockup identical to Scenario 1 while a mockup of Satellite 2 is a sponge satellite to ensure safety during impact (see Fig. 4). In this scenario, an inelastic collision is achieved during impact (the sponge damps the kinetic energy).

# 5 Experiments and Results

The two scenarios described in Section 4 are used to evaluate the usefulness of VFDM-based Cartesian Controller for on-orbit emulations. For Scenario 1, only the ceiling-mounted robot is used. For Scenario 2, both robots are utilized. As shown in Fig. 4, each robot has a mockup satellite mounted at the flange. The center-of-mass (COM) of the mockup is selected as the tool-center-point (TCP) of the robot to ensure that the robot TCP motion expressed in frame $\underline{R}$ coincides with the satellite motions expressed in $\underline{R}$. The inertial parameters of the mockups are specified in the ODS node. For convenience, the center of the $\underline{R}$ frame is considered within the limits of the ZeroG-Lab operational workspace, as indicated in Fig. 4. The robot payload is calibrated according to the weight of the mockup satellites.

The F/T sensor mounted on the robot flange measures the forces and torques applied during contact (the human operator during Scenario 1 and the mockup collision in Scenario 2). Force transformations are performed to obtain the generalized forces in the appropriate frame. These forces are delivered to the ODS ROS node along with the initial conditions of the mockups (robot's TCP position and orientation) to assign the initial state of the satellites in the ODS. Parameters, such as initial translational and rotational velocities and other scaling factors, are also specified in the ODS ROS node. In cases where the estimated impact of measured torque values, including noise from the F/T sensors, or the orientation of the satellites, is beyond the range of motion that the robots can deliver before reaching the safety stop limits, the values of torque are thus scaled down by a factor of 2000. The ODS uses the force/torque information along with the current state of the mockup to generate motion waypoints during the flight, similar to the one exercised in [21]. These waypoints serve as input set-points to the VFDM based Cartesian motion controller, and the robot TCP moves along a trajectory identical to the one of a free-floating object in space.

Table 1 lists the parameters used for the VFDM Cartesian Controller. It shows the controller gains for the VFDM-based Cartesian motion controller in Eq. (13) and the Inertial Matrix $H$ in Eq. (10). Here $m_e$ and $I_e$ are the mass and inertia of the last link of the virtual model, and $m_l$ and $I_l$ are the mass and inertia values assigned to the remaining links in the kinematic chain. The dynamics of the satellites are calculated based on a 4U CubeSat (2x2 configuration) with 1 kg mass and uniform mass density. Based on preliminary tests in the ZeroG-Lab, it was verified that large gains result in a fast and accurate IK solver. Selecting lower gains results in lower accuracy but smoother robot motion. The choice of gains in Table 1 was made accepting a trade-off between smoothness and accuracy.

**Table 1    Controller and Virtual Model Parameters**

| | |
|---|---|
| $P_{x,y,z}$, $D_{x,y,z}$ | 10.0, 0.0 |
| $P_{R_x,R_y,R_z}$, $D_{R_x,R_y,R_z}$ | 1.0, 0.0 |
| $m_e$, $m_l$ | $1\ kg$, $0.01\ kg$ |
| $I_e$ | $diag[1,1,1]\ kg \cdot m/s^2$ |
| $I_l$ | $10^{-6} I_e\ kg \cdot m/s^2$ |

## 5.1 Scenario 1: Free-Floating Satellite Motion

In this experiment, the mockup satellite experiences intermittent external forces exerted by an operator. The satellite flight path is altered due to these forces; the response is illustrated in Fig. 5 with a sequence of pictures that demonstrate the free-floating motion of the mockup.

A smooth motion in response to applied force was also observed at near-singular configurations. This highlights the VFDM-based Cartesian motion controller's capacity to maintain stability at these configurations, effectively mitigating issues like getting stuck or experiencing oscillations. The time history of force inputs imparted by the operator during the experiment along different axes is presented in Fig. 6 (top). The center and bottom plots show the change in velocity and position resulting from these external forces, respectively.

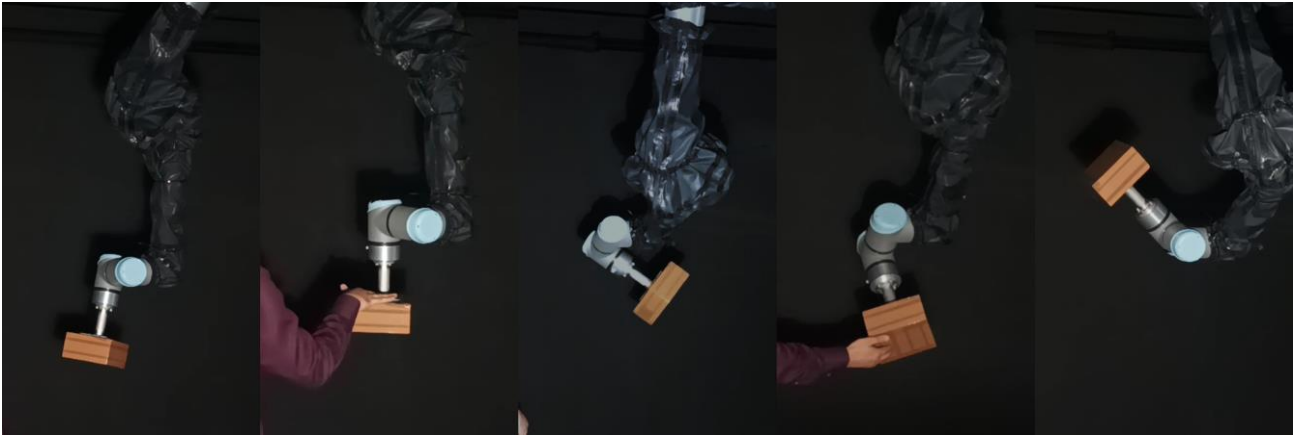

**Fig. 5** Snapshots of satellite states at different times under free floating behavior with externally applied forces by human touch.

## 5.2 Scenario 2: Free-Floating Interaction (Collision)

The simultaneous motion of two satellites caused by the force transfer during their collision is demonstrated in this experiment. The satellites start at initial states sufficiently distant from each other. An initial velocity is specified for both mockups causing them to move toward each other. The ODS generates the necessary waypoints for the VFDM-based Cartesian motion controller once the initial velocity is specified. Due to the setup, an inelastic collision occurs between the satellites during their free-floating motion. Such impact ensures safety but at a cost that the non-conservative forces caused during these experiments are not precisely reproducible. Nevertheless, the satellites' overall behavior only depends on the force values measured at the robot's TCPs.

The sequence of satellite motions throughout the experiment is shown in Fig. 7, where image (1) describes the initial states, image (2) is the collision, and the other images show the post-collision motion. Image (6) shows the final state of the mockups before the "safety limit stop" enforced by the robot's internal controller that avoids self-collision, is activated.

A force transfer occurs during the impact resulting in an exchange of equal but opposite forces on the satellites, evident in Fig. 8. These forces are expressed in the $R$ observer frame. The highest impact force occurred along $x$-axis. As the robot's motion is not strictly decoupled, lighter impact forces were detected along the other axes. The evolution of velocity and position quantities (for Satellite 2) are plotted in Figs. 9 and 10 (bottom), respectively, that corresponds to the force profile in Fig. 8 (bottom). Post impact, the direction of motion of Satellite 2 (also Satellite 1) changes, consequently the velocity and position. The ODS determines the change in motion in response to the impact generated during the collision.

A comparable consistency in the velocity profiles generated from ODS and that of the robot motion (i.e., desired and actual) is evident in Fig. 9 measured for Satellite 2. Of course, the actual readings have inherent noise. Similar adjacency in velocity profiles between simulation and robot motion tests demonstrated with a torque-controlled robot is available in literature [23]. The motion tracking accuracy of the VFDM-based Cartesian Controller, in position states, is evident from both Figs. 10 and 11. Before and after the impact, the satellite mockups track the motion generated by the ODS with errors less than

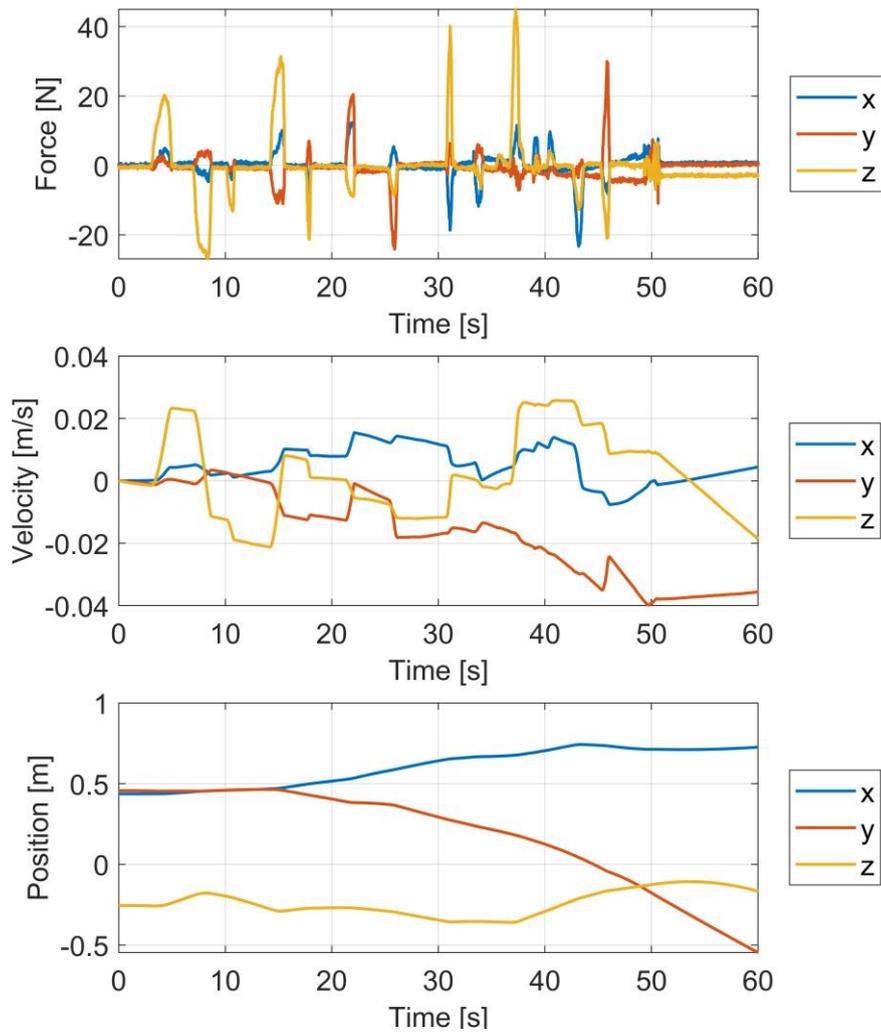

**Fig. 6** Free floating behavior of a satellite under externally applied forces by human touch. The impact of force (top) resulting in changes in velocity (center) and position (bottom) is illustrated.

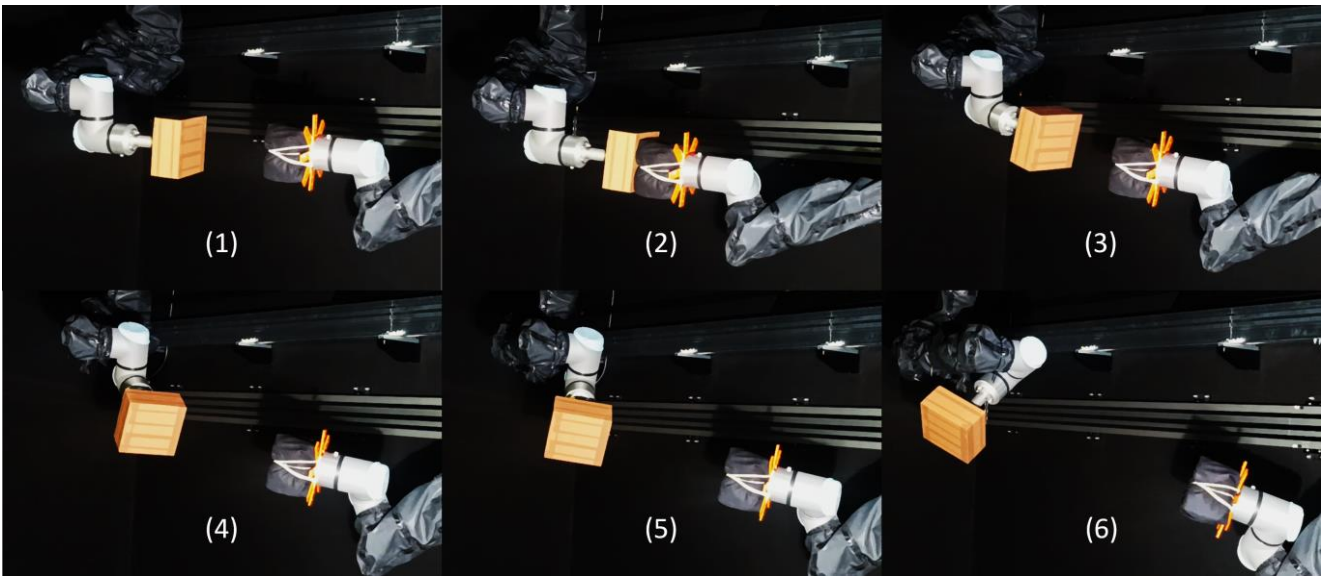

**Fig. 7** Satellite states during the process of impact. (1) Before impact, (2) During impact, (3)-(6) After impact.

0.01 m during the entire experiment. The error has a correlation with the magnitude of velocity along the waypoints. During impact, there is almost a non-zero relative motion between the two bodies (see

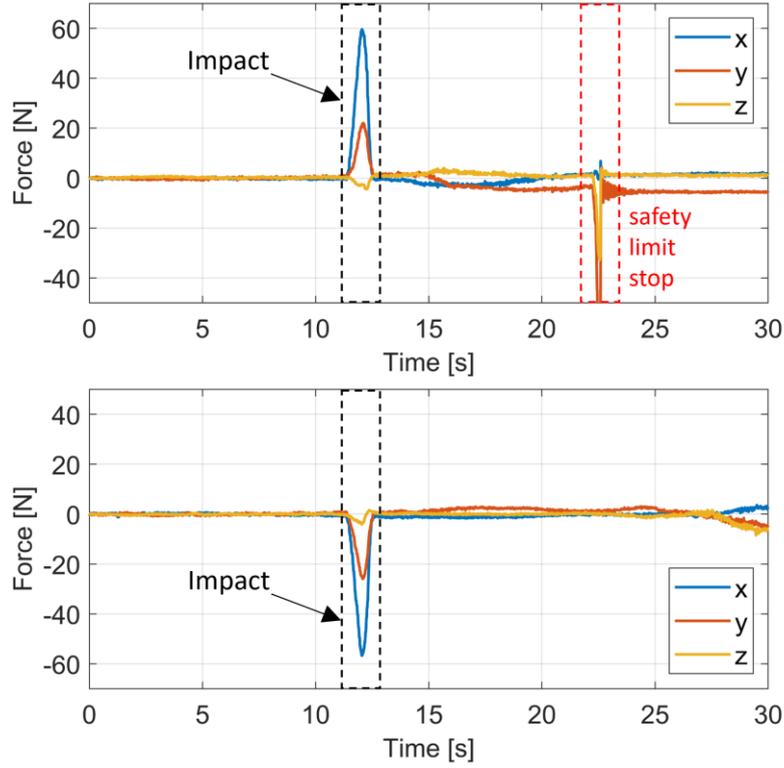

**Fig. 8** Force transfer between satellites during orbital interactions. Top: Satellite 1; bottom: Satellite 2. Force values were measured along the CW directions.

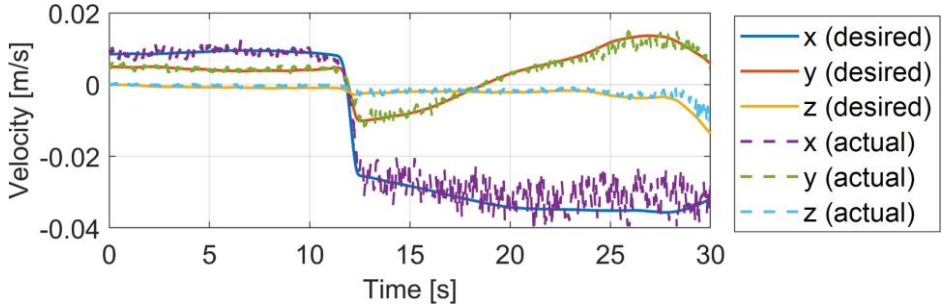

**Fig. 9** Motion Tracking Accuracy in velocity states. Comparison of the desired satellite velocities (from ODS) and actual velocities (robot motion) for Satellite 2.

Fig. 9), and more waypoints are available over a shorter distance. Subsequently, the robot tracks these waypoints more precisely, and the resulting error drops. Post-impact, when the magnitude of velocity increases, the spacing between the generated way-points is larger and therefore results in an increase in errors. With higher controller gains, the magnitude of errors can be reduced [13].

The proximity between ODS and robot-driven mockup velocity resembles the behavior observed in similar experiments performed using a torque-controlled robot [23]. The latter indicates that the proposed approach could provide an alternative to on-orbit emulation testbeds that use torque-controlled manipulators, which could be especially useful, given that position-controlled robots are lower in cost and limited in their capabilities compared to torque-controlled robots.

# 6 Conclusion

This work extends the application of Virtual Forward Dynamics Models (VFDM) for Cartesian Motion Control in robotic manipulators to emulate on-orbit scenarios within a versatile ground-based

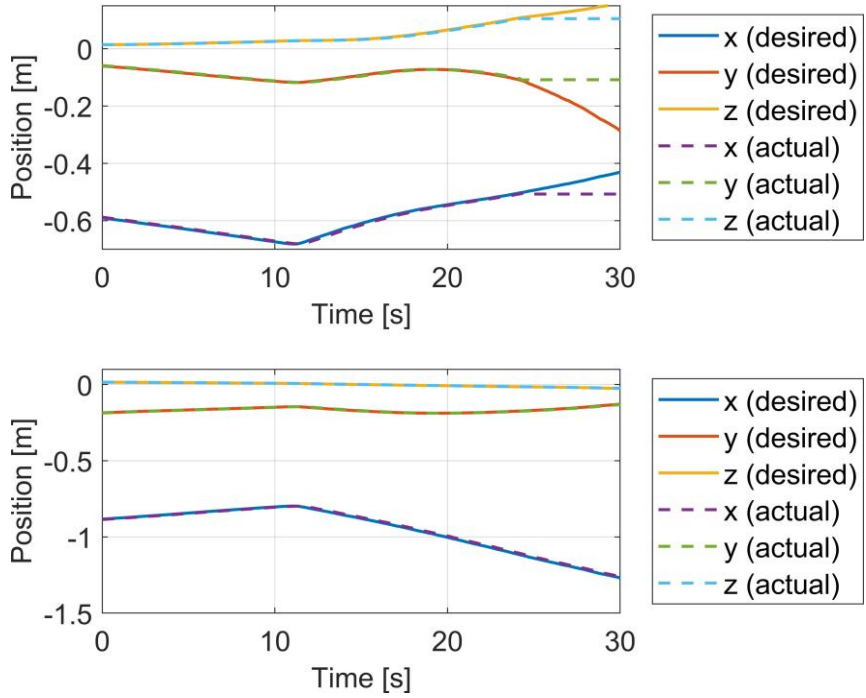

**Fig. 10** Desired satellite states (from ODS) and actual path (executed by the robot) for Satellite 1 (Top) and Satellite 2 (Bottom). The motion of Satellite 1 is restricted beyond the safety limit stop.

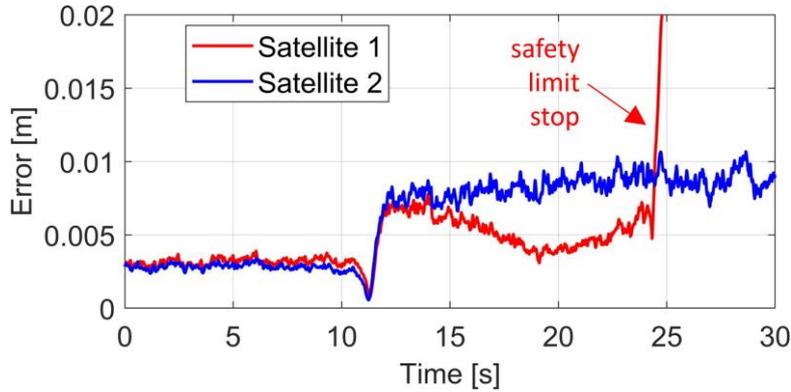

**Fig. 11** Isochronous error in satellite states between the actual executed trajectory and the desired path. The motion of Satellite 1 is restricted beyond the safety limit stop.

Hardware-in-the-Loop (HIL) testing facility. A control framework that integrates an Orbital Dynamic Simulator (based on the Clohessy-Wiltshire model) with the VFDM-based Cartesian Motion Control framework is also proposed. Two experiments were conducted; one to replicate the fundamental free-floating motion of a satellite under externally applied forces and torques, and another to simulate satellite interactions in the form of controlled collisions during free-floating motion. The changes in velocity and position profiles generated by ODS corresponding to the force/torque inputs measured on the fly establish reactive free-floating motion. A stable robot behavior is observed at near singular configurations when emulating free-floating motion. The low position tracking errors between ODS-generated profiles and the mockups validate the feasibility of VFDM for on-orbit emulation applications. Precise sensing of force and torque acting on the satellite body, rather than through moment conversion and frame transformations, can potentially enhance the fidelity of these on-orbit emulation tests. Future research will explore this idea further by integrating appropriate sensing modalities on the satellite. Furthermore, the presented concept will be extended and validated for other on-orbit operations.